\documentclass[sigconf]{acmart}

\AtBeginDocument{%
  \providecommand\BibTeX{{%
    \normalfont B\kern-0.5em{\scshape i\kern-0.25em b}\kern-0.8em\TeX}}}

\usepackage{balance}

\copyrightyear{2020}
\acmYear{2020}
\setcopyright{acmlicensed}
\acmConference[MM '20]{Proceedings of the 28th
ACM International Conference on Multimedia}{October 12--16, 2020}{Seattle,
WA, USA}
\acmBooktitle{Proceedings of the 28th ACM International Conference on
Multimedia (MM '20), October 12--16, 2020, Seattle, WA, USA}
\acmPrice{15.00}
\acmDOI{10.1145/3394171.3413691}
\acmISBN{978-1-4503-7988-5/20/10}

\begin{document}
\fancyhead{}
\title{Single-Shot Two-Pronged Detector with Rectified IoU Loss}


\author{Keyang Wang}

\affiliation{
  \institution{School of Microelectronics and Communication Engineering, Chongqing University}
 \streetaddress{Shazheng street No.174, Shapingba District}
  \city{Chongqing}
  \state{China}
  \postcode{400044}
}
\email{wangkeyang@cqu.edu.cn}

\author{Lei Zhang}
\authornote{Corresponding author}

\affiliation{
  \institution{School of Microelectronics and Communication Engineering, Chongqing University}
 \streetaddress{Shazheng street No.174, Shapingba District}
  \city{Chongqing}
  \state{China}
  \postcode{400044}
}
\email{leizhang@cqu.edu.cn}

\begin{abstract}
In the CNN based object detectors, feature pyramids are widely exploited to alleviate the problem of scale variation across object instances. These object detectors, which strengthen features via a top-down pathway and lateral connections, are mainly to enrich the semantic information of low-level features, but ignore the enhancement of high-level features. This can lead to an imbalance between different levels of features, in particular a serious lack of detailed information in the high-level features, which makes it difficult to get accurate bounding boxes. In this paper, we introduce a novel two-pronged transductive idea to explore the relationship among different layers in both backward and forward directions, which can enrich the semantic information of low-level features and detailed information of high-level features at the same time. Under the guidance of the two-pronged idea, we propose a Two-Pronged Network (TPNet) to achieve bidirectional transfer between high-level features and low-level features, which is useful for accurately detecting object at different scales. Furthermore, due to the distribution imbalance between the hard and easy samples in single-stage detectors, the gradient of localization loss is always dominated by the hard examples that have poor localization accuracy. This will enable the model to be biased toward the hard samples. So in our TPNet, an adaptive IoU based localization loss, named Rectified IoU (RIoU) loss, is proposed to rectify the gradients of each kind of samples. The Rectified IoU loss increases the gradients of examples with high IoU while suppressing the gradients of examples with low IoU, which can improve the overall localization accuracy of model. Extensive experiments demonstrate the superiority of our TPNet and RIoU loss.
\end{abstract}



\maketitle

\section{Introduction}
Along with the advances in deep convolutional networks, lots of object detectors have been developed in recent years. On the whole, all the detectors that use deep CNN can be divided into two categories: (1) The multi-stage approaches, including \cite{girshick2015fast, ren2015faster, lin2017feature, he2017mask, cai2018cascade,tan2019learning}. For the multi-stage object detectors, multi-stage classification and localization are applied sequentially, which make these models more powerful on classification and localization tasks. So these approaches have achieved the top performance on benchmark datasets. (2) The one-stage approaches, including \cite{liu2016ssd, fu2017dssd, redmon2016you, lin2017focal, zhang2018single,Nie_2019_ICCV}. The one-stage approaches apply object classifiers and regressors in a dense manner without object-based pruning. The main advantage of the one-stage detectors is their efficiency, but the detection accuracy is usually inferior to the two-stage approaches.

It is well known that enhancing the representation of features by exploiting the layer-wise correlation and dependence between different layers is an effective way to improve the performance of object detection. FPN \cite{lin2017feature} uses a top-down pathway and lateral connections to combine the low-level features and the high-level features and achieve information interaction between different layers. M2Det \cite{zhao2019m2det} presents Multi-Level Feature Pyramid Network (MLFPN) to construct feature pyramids for detecting.

In our opinion, the FPN-based detectors above only take into account one direction of the feature enhancement which may cause the imbalance between low-level and high-level features. For example, the original FPN, which strengthens feature via a top-down pathway and lateral connections, is mainly to enrich the semantic information of low-level features, but ignores the enhancement of high-level features. This can lead to an imbalance between different levels of features, in particular a serious lack of detailed information in the high-level features, which makes it difficult to get accurate bounding boxes although the network can regress some rough boundaries.

\begin{figure}[h]
  \centering
  \includegraphics[width=6.6cm,height=5.9cm]{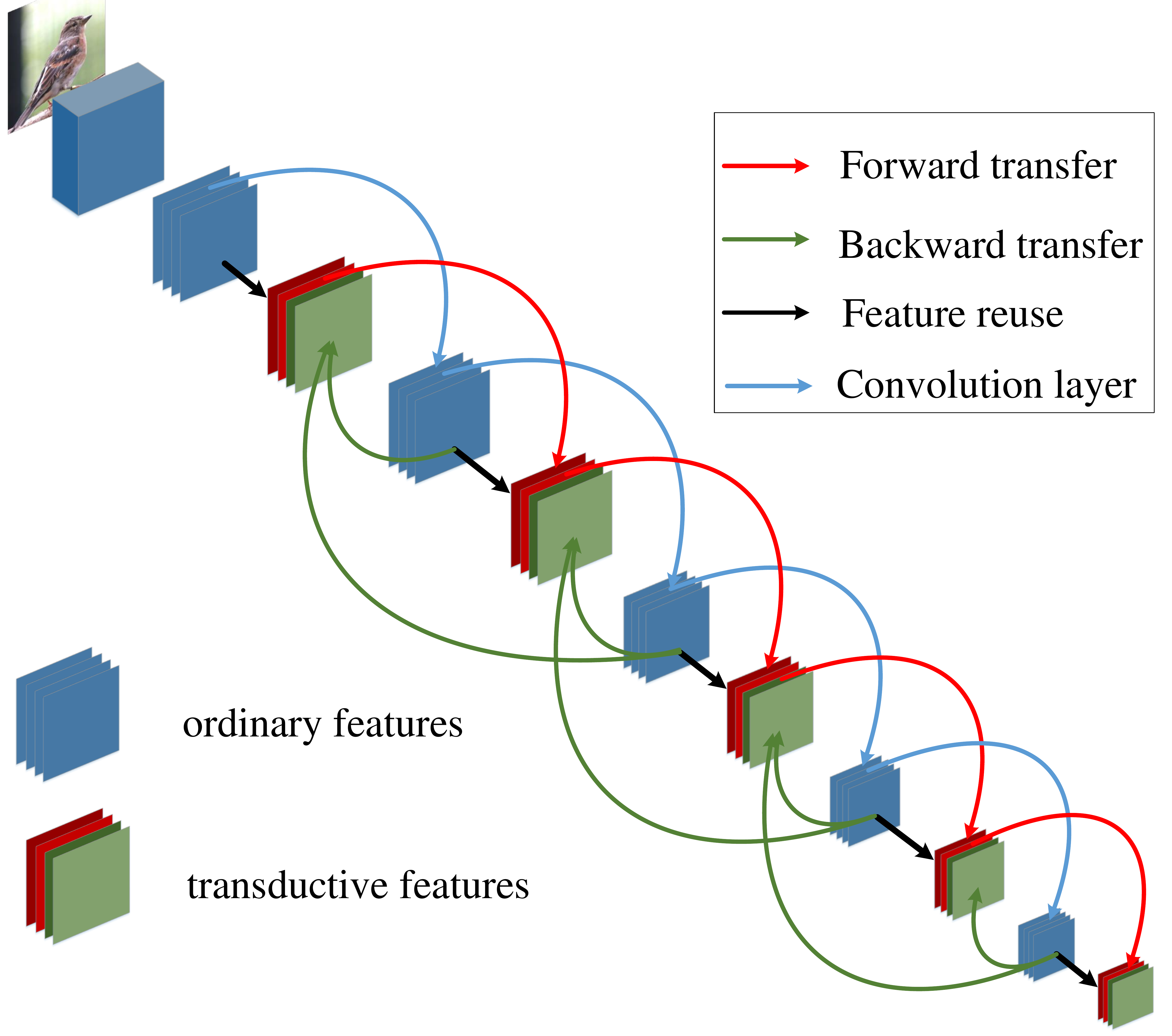}
  \vspace{-0.2cm}
  \caption{Our proposed two-pronged transductive structure, in which the backward transfer (shown as the green lines with arrows) achieves the feature transfer from high-level layers to low-level layers, and the forward transfer (shown as the red lines with arrows) achieves the transfer from transductive features to high-level features.}
  \label{motivation}
\vspace{-0.5cm}
\end{figure}

In order to alleviate this problem, we think that high-level features and low-level features should be equally interacted, which means that each layer needs to receive more abstract information from its upper layer and meanwhile get more basic cues or detailed information from its lower layer, and we call this two-pronged effect. In this paper, we originally introduce a novel two-pronged transductive idea to explore the relationship and two-pronged effect among different layers in both backward and forward directions, which can improve the classification performance in low-level layers and the regression performance in high-level layers. Following the two-pronged idea, we propose a Single-Shot Two-Pronged Network (TPNet) consisting of multiple two-pronged layer-wise interaction blocks named two-pronged Transductive (T) blocks to achieve bidirectional transfer between high-level features and low-level features. As shown in Figure~\ref{motivation}, five T blocks are added after the backbone. In each T block, we first transfer multiple high-level features of low-resolution but stronger semantics to the low-level features of high-resolution but weak semantic by the backward transfer layers, as shown by the green lines with arrows. We gain rich semantic information at all levels by the multi-level fusion, which is helpful for object classification task in low-level layers. Secondly, in order to compensate for the detailed information loss in down-sampling process of network and get more basic cues from lower layer, the forward transfer layers are designed to propagate the enhanced features to higher-level layers, as shown by the red solid lines with arrows. We obtain more detailed information in deeper layers with the forward transfer layer, which can be effective for accurate object location. Through several T blocks, the relationship and two-pronged effect among different layers can be explored after training and different level features can be jointly interacted to enhance the representation ability on both sides.

In the single-shot detectors, the matching strategy based on IoU is often used for preparing the training samples. However, this IoU based strategy neglects a fact that the IoU distribution of all the samples is seriously imbalanced. This will cause that the gradient of localization loss (e.g., smooth $\mathcal{L}_1$ loss, IoU loss) for single-shot detectors are dominated by outliers (low IoU levels) during training phase, which enables the detection model to be biased toward the outliers. The issue always leads to the network unable to regress accurate bounding boxes. So if we can rectify the gradients of each kind of examples during training phase, the adverse effects caused by this issue will be mitigated well. Motivated by this, we propose an adaptive IoU based localization loss, named Rectified IoU (RIoU) loss. Our Rectified IoU loss up-weights the gradients of examples with high IoU while suppressing the gradients of examples with low IoU. Training with Rectified IoU loss, the huge amount of cumulated gradient produced by easy examples (high IoU levels) can be up-weighted and the outliers (low IoU levels) can be relatively down-weighted. In the end, the contribution of each kind of example will be balanced and the training can be more efficient and stable.

With the Two-Pronged Transductive (T) blocks for feature learning and the Rectified IoU loss for training, our TPNet can regress more accurate bounding boxes in object detection. In order to evaluate the effectiveness of the proposed TPNet, we apply it on the very challenging PASCAL VOC \cite{everingham2007pascal} and MS COCO \cite{lin2014microsoft} benchmarks. The detection results demonstrate the competitiveness of our TPNet over state-of-the-arts, especially in the case of higher IoU threshold. In summary, this paper makes three main contributions:

\hangafter=1
\setlength{\hangindent}{2.5em}
1. We propose a novel two-pronged transductive idea to explore the two-pronged effect between high-level features and low-level features through the forward and backward feature transfer. And under the guidance of the two-pronged transductive idea, we propose a novel Two-Pronged Network (TPNet) consisting of multiple two-pronged layer-wise transfer blocks named Two-Pronged Transductive (T) blocks, which can achieve more accurate bounding boxes regression.

\hangafter=1
\setlength{\hangindent}{2.5em}
2. We introduce an IoU based localization loss, named Rectified IoU (RIoU) loss, to rectify the gradients of each kind of examples, which can prevent the gradients of localization loss from being dominated by outliers during training phase and ensure accurate bounding box regression ability of the whole detector.

\hangafter=1
\setlength{\hangindent}{2.5em}
3. With two-pronged Transductive (T) blocks and the Rectified IoU loss, a scale-aware object detector (TPNet) is materialized. The proposed T block and RIoU loss can be easily plugged and played in most existing detectors. The proposed TPNet achieves competitive results on the PASCAL VOC and MS COCO benchmarks.

\begin{figure*}
\begin{center}
  \centering
  \includegraphics[width=14.0cm,height=8.5cm]{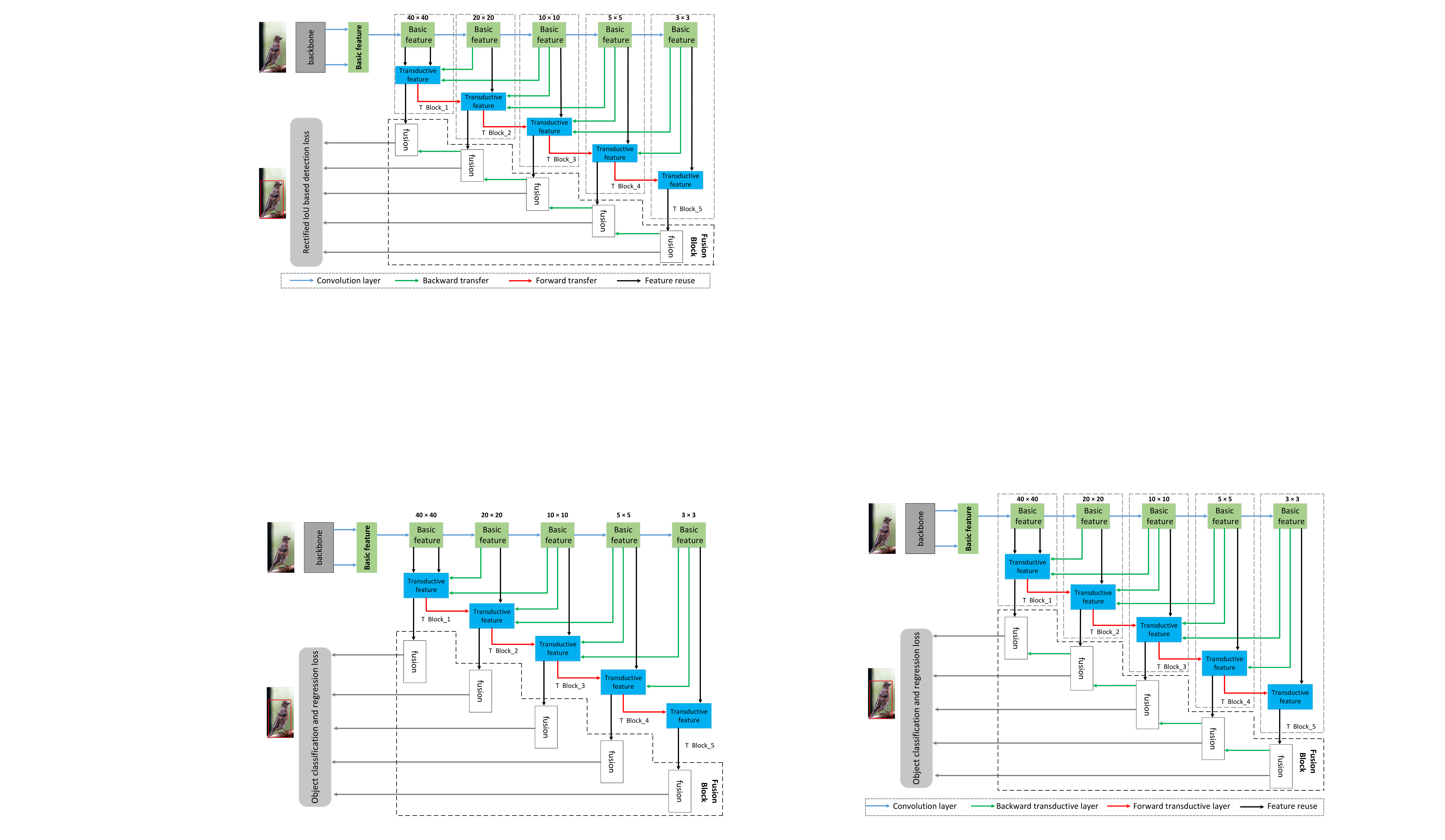}
\end{center}
\vspace{-0.4cm}
   \caption{The overall architecture of TPNet which includes several T blocks and a Fusion block. The T blocks achieve
bidirectional feature transfer. The Fusion block further constrain the features of each T block.}
\label{network}
\vspace{-0.2cm}
\end{figure*}
\section{RELATED WORK}
\textbf{Model architectures for object detection.} Benefited from the power of Deep ConvNets, CNN has achieved great success in the object detection field. All the CNN-based detectors can be roughly divided into two categories, \emph{i.e.,} two-stage detector and one-stage detector. The two-stage detector consists of two parts. The first part is responsible for generating a set of candidate object proposals, e.g., Selective Search \cite{uijlings2013selective}, EdgeBoxes \cite{zitnick2014edge}, RPN \cite{ren2015faster}. The second part determines the accurate object regions and the corresponding class labels using convolutional networks according to the candidate object proposals. Its descendants (e.g., R-CNN \cite{girshick2015region}, Fast R-CNN \cite{girshick2015fast}, Faster R-CNN \cite{ren2015faster}, R-FCN \cite{dai2016r}, FPN \cite{lin2017feature}, Mask RCNN \cite{he2017mask}, CoupleNet \cite{zhu2017couplenet}) achieve dominant performance on several challenging datasets. However, the most serious limitation of the two-stage approaches is inefficiency. In view of high efficiency, the one-stage approaches attract much more attention recently. YOLO \cite{redmon2016you} divides the input image into many grids and uses a single forward convolutional network to perform localization and classification on each part of the image, which is very fast. SSD \cite{liu2016ssd} is another efficient one-stage detector, which adds a series of progressively smaller convolutional layers to generate pyramid feature maps and two 3 $\times$ 3 convolutional layers to predict the class scores and location offsets of the default bounding boxes. After that, more advanced one-stage detectors (e.g., RetinaNet \cite{lin2017focal}, RefineDet \cite{zhang2018single}, RFB \cite{liu2018receptive}, M2Det \cite{zhao2019m2det}) even achieve higher accuracy than some two-stage detectors. Especially, in order to solve the problem that the detection result is too sensitive to the size of the anchor and simultaneously avoid the complex IoU computation and matching between anchor boxes and ground-truth boxes during training, some anchor-free detectors are proposed, including CornerNet \cite{law2018cornernet}, FCOS \cite{tian2019fcos}, CenterNet \cite{duan2019centernet}.

\textbf{IoU based localization loss.} The $\mathcal{L}_n$-norm loss functions are usually adopted in bounding box regression. But as $\mathcal{L}_n$-norm loss has proved to be not tailored to the regression task because of their scale-sensitivity, some IoU-based regression losses are proposed.  UnitBox \cite{yu2016unitbox} adopts the IoU loss instead of $\mathcal{L}_1$-norm loss to regress the bounding boxes (bbox) and achieves more accurate location in face detection. GIoU \cite{rezatofighi2019generalized} is a generalized version to address the weakness of IoU loss by extending the concept to non-overlapping cases. DIoU \cite{zheng2020distance} incorporates the normalized distance between the predicted bbox and the GT bbox on the basis of GIoU. The IoU-based losses have received increased attention due to their effectiveness under a high IoU threshold.

\textbf{Gradient rectified localization loss functions.} In order to balance the contribution of each kind of examples and prevent the gradients of localization loss from being dominated by outliers, some detectors have also made some explorations on gradient rectified localization loss. GHM \cite{li2019gradient} analyzes the example imbalance in one-stage detectors in term of gradient norm distribution and propose a gradient harmonized localization loss GHM-R. Libra R-CNN \cite{pang2019libra} claims that the overall gradient of smooth $\mathcal{L}_1$ loss is dominated by the outliers when balancing classification and localization task directly. As a result, the balanced $\mathcal{L}_1$ loss is proposed to increase the gradient of easy examples and keep the gradient of outliers unchanged. However both the GHM-R and balanced $\mathcal{L}_1$ loss are based on smooth $\mathcal{L}_1$ loss, which is not invariant to the scale.
Different from that, our proposed Rectified IoU loss is an IoU based gradient rectified localization loss functions. This means that our Rectified IoU loss can automatically rectify the gradients of different samples while ensuring the scale invariance.

\section{Two-Pronged Network}
\begin{figure*}
\begin{center}
  \centering
  \includegraphics[width=13cm,height=5cm]{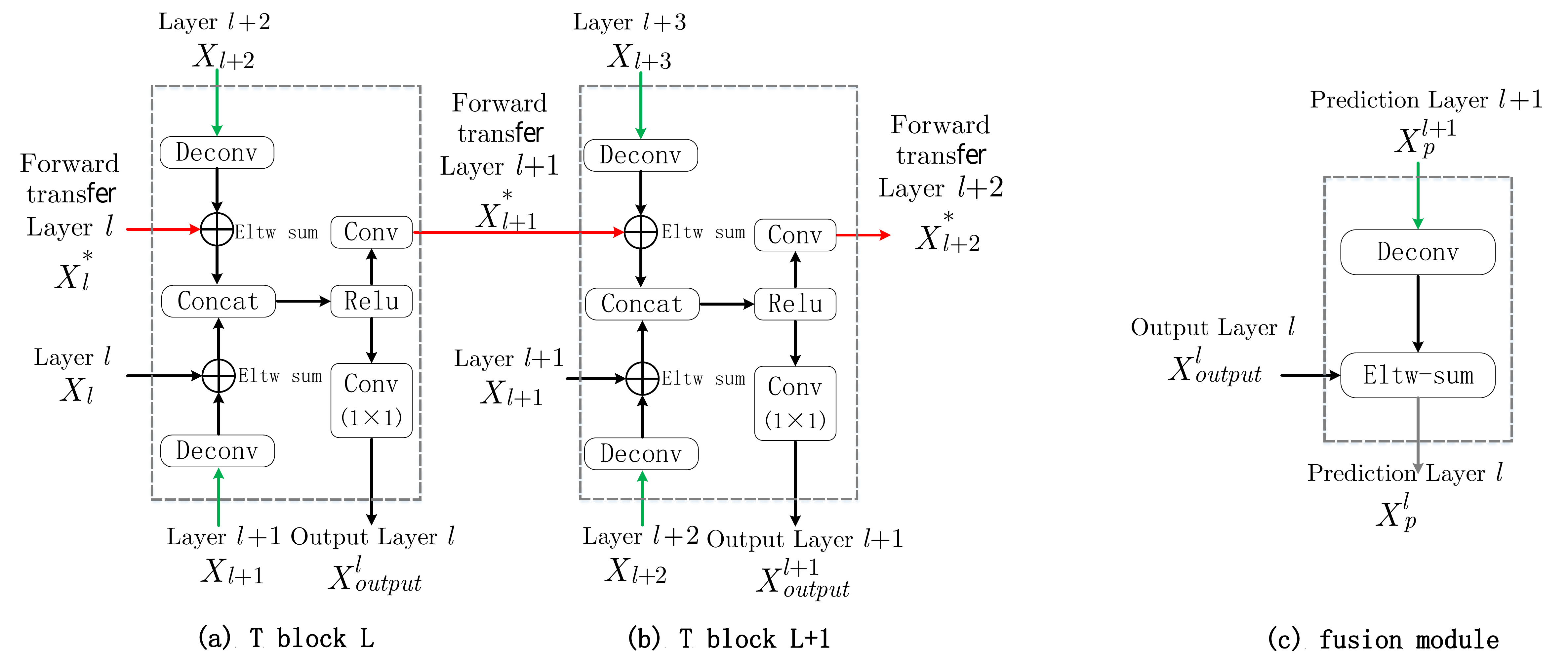}
\end{center}
\vspace{-0.3cm}
   \caption{Structural details of two blocks. (a) Two-Pronged Transductive (T) block L, (b) Two-Pronged Transductive (T) block L+1: the
core module of the TPNet which achieves multi-neighbor backward transfer and forward transfer. (c) fusion module: the basic module of
Fusion block.}
\label{tblock}
\vspace{-0.2cm}
\end{figure*}

Under the guidance of the two-pronged idea, we propose a novel Two-Pronged Network (TPNet). The overall architecture of the TPNet is shown in Figure~\ref{network}. We first feed the image into backbone (e.g., ResNet-50) to extract the basic features. Then, the basic features are fed into the multiple T blocks to extract more representative features by bidirectional transfer. Lastly, we construct a feature pyramid for the final object detection. Similar to SSD, we produce dense bounding boxes and category scores on the feature pyramid, followed by the non-maximum suppression (NMS) operation to produce the final results. In this section, we first introduce in detail the architecture of our TPNet consisting of multiple two-pronged layer-wise transfer blocks named Two-Pronged Transductive (T) blocks in section~\ref{T_block} and Fusion (F) block in section~\ref{F_block}. Then we describe the details of Rectified IoU (RIoU) loss in section~\ref{RIoU_loss}.
\vspace{-0.0cm}
\subsection{Two-Pronged Transductive (T) Blocks}
\label{T_block}
In order to introduce the process of the transfer between different T blocks, we present the detailed structure and
connection of two neighbor T blocks, as shown in Figure~\ref{tblock} (a), (b). We can clearly find that our T block is mainly divided into two parts, i.e., multi-neighbor backward transfer and forward transfer.

\textbf{Multi-neighbor backward transfer.} As we know, the shallow layer (e.g., conv2\_x for ResNet50) in the network is useful for small object detection because of its abundant detailed information. However, due to the lack of semantic information, those shallow layers perform poorly on classification task. So in order to improve the classification performance of shallow layers, we transfer multiple high-level features to the low-level features by backward transfer layers. This will help low-level feature maps acquire different semantic information from different deeper layers,
which can improve the performance of the classifier in the low-level layer. Unlike DSSD \cite{fu2017dssd} which does up-sampling and fusion from the last level one by one, we fuse the current features with the features of its multi-neighbor layers to gain the transductive features. For example, if the conv2\_x in ResNet-50 is the current layer, then the conv4\_x in ResNet-50 is the first neighbor layer, while the basic feature layer conv5\_x which we build is the second neighbor layer and so on. The specific structure may refer to Figure~\ref{tblock} (a) and (b) which adopt 2-neighbor layer transfer. Assuming $X_l$ , $X_{l+1}$, $X_{l+2}$ to be the feature maps of the layer $l$, layer $l + 1$ and layer $l + 2$, and $X_{out}^l$ to be the transductive features after multi-neighbor backward transfer. So the features of the $l^{th}$ aggregation layer can be described as follows:
\vspace{-0.0cm}
\begin{equation}
\label{eq1}
X_{out}^l=Relu((X_l \oplus X_{l+1}^{\uparrow}) \uplus (X_l^* \oplus X_{l+2}^{\uparrow}))
\end{equation}
where ``$\uparrow$" is the up-sampling operation function, ``$\oplus$" is the element-wise sum operation of feature maps,``$\uplus$" is the concatenate operation between feature maps, $Relu(\cdot)$ is the activation operation and $X_l^*$ is the feature map of the $l^{th}$ forward transfer layer, which will be introduced later. This part achieves the transfer from high-level layers to low-level layers by the up-sampling and concatenated operation, which is effective for small object detection in low-level layers.

\textbf{Forward transfer.} In previous methods, such as FPN \cite{lin2017feature} and DSSD \cite{fu2017dssd}, the features of the aggregation layers, which are gained by constructing a feature pyramid, are only fed into the classifier and regressor but without making any contributions to the high-level layers. We think those aggregation layers with abundant detailed information are useful for the accurate bounding box regression in high-level layers where much detailed information is lost in the down-sampling process. So in order to achieve more accurate bounding box regression, the forward transfer layers are proposed to compensate for the detailed information loss in down-sampling process and enable upper layers with more basic cues from lower layers. In fact, it is a concise but very effective convolution layer as shown by the red solid lines with arrows in Figure~\ref{tblock} (a) and (b). Suppose $X_{out}^l$ to be the aggregation output of multi-neighbor backward transfer, and in order to transfer $X_{out}^l$ to the next T block, a learnable convolution layer is used to obtain the forward transductive features $X_{l+1}^*$ of the next T block. The forward transductive features will be fused with the second neighbor layer of the next T block to achieve forward feature transfer. The forward transductive features $X_{l+1}^*$ can be described as follows:
\begin{equation}
\label{eq2}
X_{l+1}^*=Conv(X_{out}^l)
\end{equation}

From the Eq.(\ref{eq1}) and Eq.(\ref{eq2}), the T blocks can be divided into the following core components: (1) We transfer the features of the first neighbor layer $X_{l+1}$ to the features of the current layer $X_l$ after deconvolution operation to obtain the $k_1$ aggregation features. (2) We transfer the features of the second neighbor layer $X_{l+2}$ to the forward transfer layer $X_l^*$ after deconvolution operation to obtain the $k_2$ aggregation features. (3) Concatenating the features of the $k_1$ and $k_2$ aggregation layers to obtain the transductive features $X_{out}^l$. (4) We design the effective forward transfer layers to propagate the enhanced features $X_{out}^l$ to high-level layers and apply a $1\times1$ convolution layer to gain the augmented features $X_{output}^l$  of the $l^{th}$ T block at the same time. By designing the T blocks, we achieve bidirectional feature transfer in a single network and generate better features at all levels.
\subsection{Fusion (F) Block}
\label{F_block}
In order to further constrain the features of each layer, we design a more concise Fusion (F) block, which consists of several simple fusion modules shown as Figure~\ref{tblock} (c), to construct a feature pyramid for the final object detection. As shown in Figure~\ref{network}, after obtaining the transductive features from multiple T blocks, then we construct a feature pyramid to further constrain the features of each layer, which includes several deconvolution and element-wise sum operations. Finally, we feed the features of the feature pyramid into the classifier and regressor for object detection.
\subsection{Rectified IoU loss}
\label{RIoU_loss}
Intersection over Union (IoU) is a popular metric in object detection, defined as:
\begin{equation}
\label{eq3}
IoU=\frac{|B_p \cap B_g|}{|B_p \cup B_g|}
\end{equation}
where $B_p$ is the predicted bbox, $B_g$ is the ground-truth. As suggested in \cite{yu2016unitbox}, the $\mathcal{L}_n$-norm loss function is not a suitable choice to obtain the optimal IoU metric, so IoU loss, defined as Eq.(\ref{eq4}), is suggested to be adopted for improving the IoU metric and regressing accurate bounding boxes.
\begin{equation}
\label{eq4}
\mathcal{L}_{IoU}=1-IoU
\end{equation}
The absolute value $|gradients(IoU)|$ of the gradients of the IoU loss can be expressed as:
\begin{equation}
\label{eq5}
|gradients(IoU)|=|\frac{\partial \mathcal{L}_{IoU}}{\partial IoU}|=1
\end{equation}
The $|gradients(IoU)|$ is the constant 1 for both easy samples (samples at high IoU levels) and hard samples (samples at low IoU levels). But the IoU distribution of all the samples is seriously imbalanced, generally, the number of samples at low IoU levels is larger than that at high IoU levels. Needless to say, the hard samples at low IoUs dominate the gradients of the location loss in the training phase, which enables the detection model to be biased toward the hard samples. So we want to explore a new IoU loss which can rectify the gradients of each kind of example. Specifically, the loss can up-weight the gradients of examples with high IoU while suppressing the gradients of examples with low IoU.

\begin{figure}[h]
  \centering
  \includegraphics[width=0.78\linewidth,height=5.5cm]{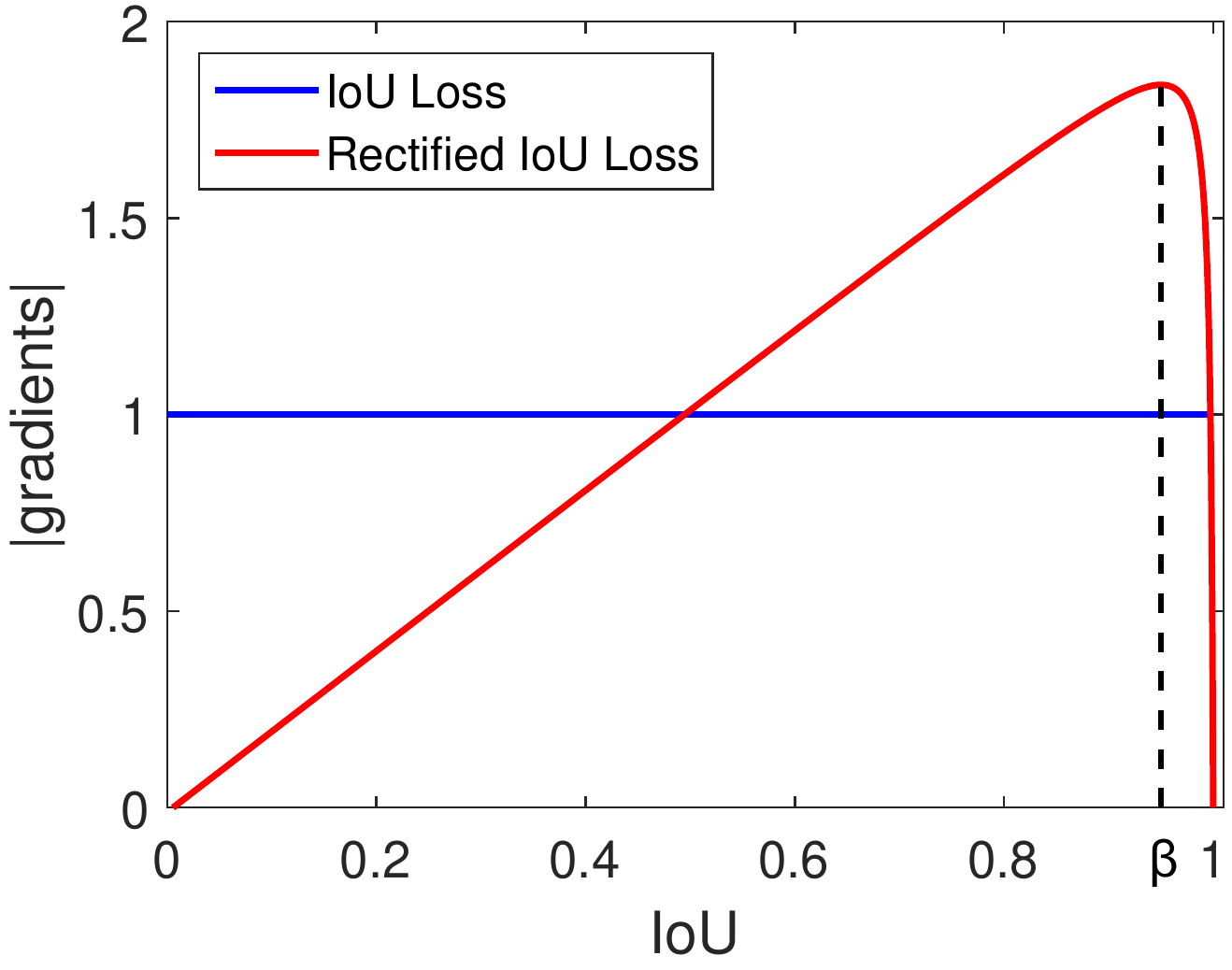}
  \vspace{-0.3cm}
  \caption{The gradients of the standard IoU loss and our proposed Rectified IoU loss. $\beta$ is the position of the inflection point.}
  \label{gradients}
  \vspace{-0.3cm}
\end{figure}

But if we always up-weight the gradients of the localization loss as the IoU increases, we will face another problem that the gradient will continue to increase when the regression is perfect ($IoU \rightarrow 1$). This means we will get the maximum gradient when two bboxes overlay perfectly ($IoU=1$), which is very unreasonable.

Combining the above two points, we proposed a Rectified IoU (RIoU) loss, whose gradient is a hyperbolic function, as shown in Figure~\ref{gradients}. The hyperbolic gradient formulation can be defined as:
\begin{equation}
\label{eq6}
|gradients(IoU)|=|\frac{\partial \mathcal{L}_{RIoU}}{\partial IoU}|=(aIoU+b)+\frac{k}{(IoU-c)}
\vspace{-0.0cm}
\end{equation}
where $a$, $b$, $c$ and $k$ are four parameters used to control the shape of the curve of the hyperbolic gradient. From the Figure~\ref{gradients}, we can clearly find that the gradient value rises first and then drops sharply as the IoU value increases. And the gradient value has an inflection point when $IoU=\beta$. In this paper, we set the $\beta$ to 0.95. This inflection point can be controlled by parameters $a$, $b$, $c$ and $k$. Conversely, we can calculate the values of the parameters according to the starting point, the ending point and inflection point of the gradient curve.

By integrating the gradient formulation defined as Eq.(\ref{eq6}) and combining the properties of IoU based localization loss, we can get the Rectified IoU loss as follows:
\begin{equation}
\label{eq7}
\mathcal{L}_{RIoU}=1-(\frac{a}{2}IoU^2+bIoU+kln|IoU-c|+t)
\end{equation}

\begin{figure}[h]
  \centering
    \vspace{-0.3cm}
  \includegraphics[width=0.8\linewidth,height=5.4cm]{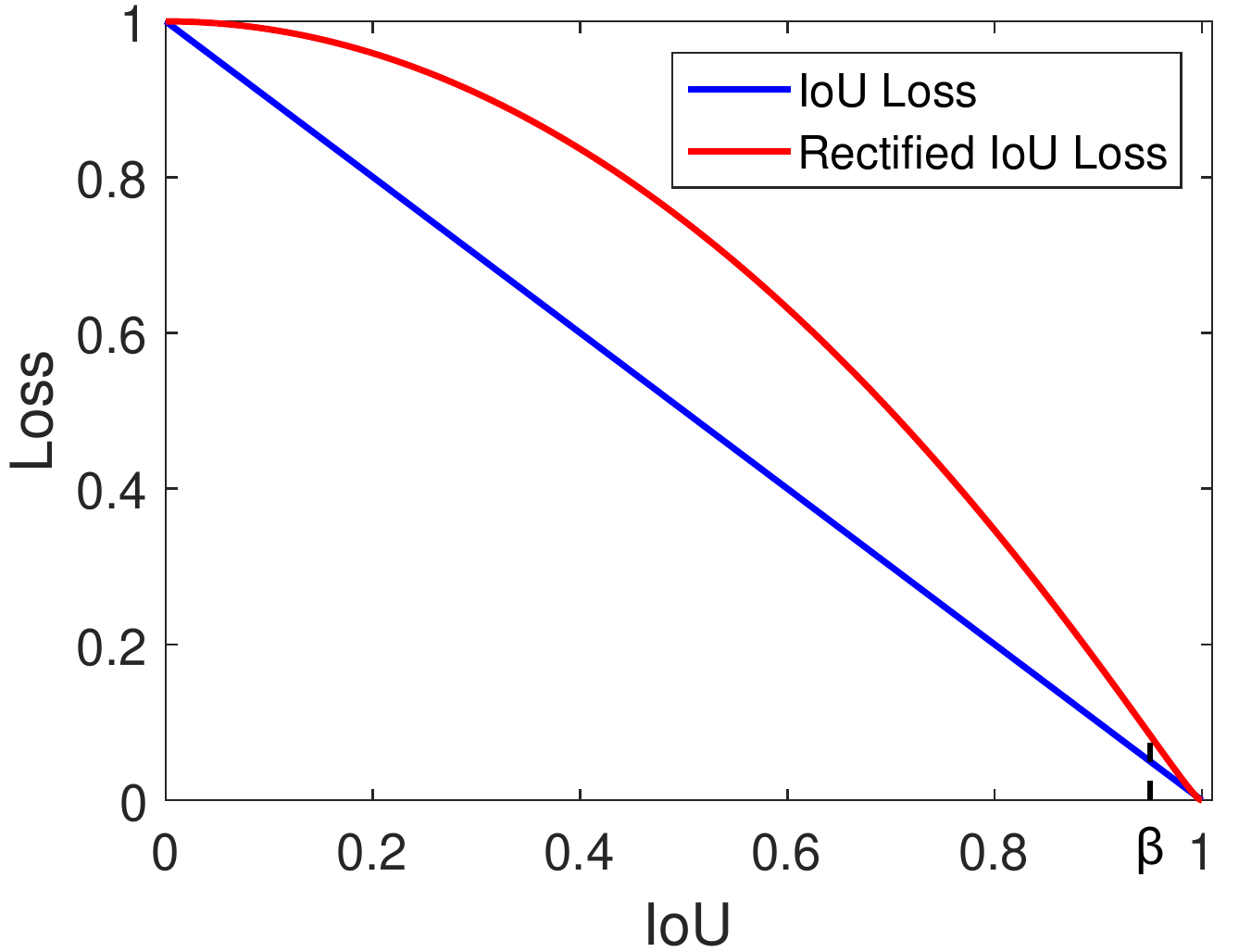}
  \vspace{-0.3cm}
  \caption{The distribution curves of the standard IoU loss and our Rectified IoU loss.}
  \label{loss}
  \vspace{-0.3cm}
\end{figure}

\textbf{Analytical Solution of the Parameters.} As the five parameters in Eq.(\ref{eq7}) are analytically determined parameters, rather than hyperparameters, so in the following, we further introduce how to calculate the values of the five parameters in Eq.(\ref{eq7}). Firstly, if we set the starting point, the ending point and the inflection point of the gradient curve as (0,0), (1,0) and $IoU=\beta$ ($\beta=0.95$ in this paper) respectively, as shown in Figure~\ref{gradients}. By substituting them into Eq.(\ref{eq6}), we can obtain three constraint equations as follows:
\begin{equation}
\label{eq8}
\left\{
\begin{aligned}
& b-\frac{k}{c}=0 \\
& a+b+\frac{k}{1-c}=0 \\
& c-\sqrt{\frac{k}{a}}=\beta
\end{aligned}
\right.
\end{equation}
Secondly, according to the two properties of IoU based localization loss: (1) The RIoU loss will be 1 when $IoU=0$. (2)The RIoU loss will be 0 when $IoU=1$, by substituting them into Eq.(\ref{eq7}), we can get another two constraint equations as follows:
\begin{equation}
\label{eq9}
\left\{
\begin{aligned}
& 1-kln|c|-t=1 \\
& 1-\frac{a}{2}-b-kln|1-c|-t=0 \\
\end{aligned}
\right.
\end{equation}
By combining the Eq.(\ref{eq8}) and Eq.(\ref{eq9}), we can calculate the values of the five parameters $a,b,c,k,t$ and get the final Rectified IoU loss formulation according to the value of $\beta$ that we set in advance. The proposed RIoU loss ($\beta$=0.95) is visualized in Figure~\ref{loss}. We can find that the distribution of RIoU loss shows an upwards convex shape when IoU$\textless\beta$, and a sunken shape when IoU$\textgreater\beta$.


\textbf{Detection Loss.}
In order to accelerate the convergence of the RIoU localization loss, inspired by the DIoU loss \cite{zheng2020distance}, we add the normalized distance between the central points of two bounding boxes to the localization loss, which can directly minimize the distance between predicted bbox and GT bbox for achieving faster convergence. With the proposed localization loss and the basic object classification loss, the overall objective of our TPNet is summarized as follows:
\begin{equation}
\label{eq11}
\mathcal{L}_{det}=\mathcal{L}_{cls}+\mathcal{L}_{RIoU}+\rho (B_p^{ct}, B_g^{ct})
\end{equation}
where $\mathcal{L}_{cls}$ is the classification loss, which is the cross entropy loss in our experiments, the $\rho(\cdot)$ is the distance function, which is smooth $\mathcal{L}_1$ in this paper. $B_p^{ct}$ is the central points of the predicted bbox, $B_g^{ct}$ is the central points of the ground-truth.

\section{Experiments}
In this section, we first conduct experiments on two widely used benchmarks for the object detection task, i.e., PASCAL VOC and MS COCO. We then conduct ablation analysis of the proposed T blocks, F block and RIoU loss in our TPNet. All of our models are trained under the PyTorch framework with SGD solver on NVIDIA Titan Xp GPUs.
\vspace{-0.1cm}
\subsection{Experiments on PASCAL VOC}
PASCAL VOC is a dataset of 20 classes for extensively evaluating the object detection algorithms. In our experiments, all models are trained on the union of the VOC 2007 \texttt{trainval} and VOC 2012 \texttt{trainval} datasets, and tested on the VOC 2007 \texttt{test} set. And the ResNet-50 is adopted as the backbone. We set the batch size as 32 for PASCAL VOC datasets. The momentum is fixed to 0.9 and the weight decay is set to 0.0005, which is consistent with the original SSD settings. We start the learning rate with ${10}^{-3}$ for 150 epochs and decay it to $10^{-4}$ and $10^{-5}$ for another 50 and 50 epochs respectively in PASCAL VOC datasets. In order to prove that our model can regress more accurate bounding boxes, a stricter COCO-style Average Precision (averaged AP at IoUs from 0.5 to 0.9 with an interval of 0.1) metrics is adopted on the PASCAL VOC dataset.
\begin{table*}
\vspace{-0.4cm}
  \caption{Comparison of detection methods on the PASCAL VOC dataset. For PASCAL VOC 2007, all methods are trained on VOC 2007 and VOC 2012 \texttt{trainval} sets and tested on VOC 2007 \texttt{test} set.}
  \vspace{-0.2cm}
  \label{sample-table1}
  \centering
  \begin{tabular}{c|c|c|p{0.7cm}p{0.7cm}p{0.7cm}p{0.7cm}p{0.7cm}p{0.7cm}}
    \toprule
    Method      &     Backbone    &    Input size     &   AP& AP$_{50}$ & AP$_{60}$ & AP$_{70}$ & AP$_{80}$ & AP$_{90}$   \\
    \hline
    \footnotesize\emph{two-stage:}& & &&&&&&\\
    Faster R-CNN \cite{ren2015faster}              & ResNet-50-FPN  & $\sim$1000$\times$600 &52.9&79.8&75.0&61.7&39.0&8.8  \\
    Cascade R-CNN \cite{cai2018cascade}            & ResNet-50-FPN  & $\sim$1000$\times$600 &58.5&80.0&74.7 &65.8&50.5& 21.5    \\
    \hline
    \hline
    \footnotesize\emph{one-stage:}& & &&&\\
    SSD300 \cite{liu2016ssd}             & VGG-16  & 300$\times$300 &52.7&77.6&72.7 &61.0&40.9&11.4  \\
    YOLOv2 \cite{redmon2017yolo9000}             & Darknet-19       &544$\times$544 &53.7&78.6&73.6 &62.0&41.6&12.8 \\
    DSSD320 \cite{fu2017dssd}             & ResNet-50    &321$\times$321 &56.1&79.6&74.8 &64.1&46.1&16.0 \\
    GIoU \cite{rezatofighi2019generalized}              & ResNet-50-FPN      &300$\times$300 &55.3&78.4&74.1 &63.5&45.9&14.6 \\
    DIoU \cite{zheng2020distance}            & ResNet-50-FPN       &300$\times$300 &55.8&78.9&74.6 &64.0&46.2&15.5 \\
    RefineDet320 \cite{zhang2018single} & VGG-16          &320$\times$320 &54.7&80.0&74.2 &63.5&43.3& 12.2\\
    DAFS320 \cite{li2019dynamic}              & ResNet-101      &320$\times$320 &58.7 &\textbf{81.0} &76.3& \textbf{66.9} &49.2 &20.0 \\
    \textbf{TPNet320}(Ours)               & ResNet-50  & 320$\times$320 &\textbf{59.4}&80.3&\textbf{76.3}&66.8&\textbf{50.9}&\textbf{22.5} \\
    \hline
    SSD512 \cite{liu2016ssd}              & VGG-16      & 512$\times$512 &57.5&79.8&76.6&66.7 &49.4&15.2 \\
    DSSD512 \cite{fu2017dssd}              & ResNet-50      &513$\times$513 &58.5&81.5&77.7&67.6 &50.0&15.8\\
    RefineDet512 \cite{zhang2018single} & VGG-16           &512$\times$512 &58.4&81.8&77.8& 67.2&49.6& 15.6            \\
    RetinaNet \cite{lin2017focal} & ResNet-101-FPN           & $\sim$1000$\times$600&59.3&81.1&77.2&67.5  &50.4&20.1            \\
    DAFS512 \cite{li2019dynamic}              & VGG-16      &512$\times$512 &59.4&\textbf{82.4}&\textbf{78.2}&67.6 &50.9&18.0 \\
    \textbf{TPNet512}(Ours)           & ResNet-50      & 512$\times$512&\textbf{61.2} &81.7&78.0 &\textbf{69.3}&\textbf{53.0}&\textbf{24.0}   \\
    \bottomrule
  \end{tabular}
\end{table*}

We compare our method with the state-of-the-art detectors on VOC 2007 \texttt{test} set. As shown in Table~\ref{sample-table1}. Without bells and whistles, our TPNet can achieve 59.4$\%$ AP when the input size is 320, which can improve AP by 3.3$\%$ compared with the baselines DSSD320 (from 56.1$\%$ to 59.4$\%$). Especially, the improvement for AP at higher IoU threshold (0.8, 0.9) is 4.8$\%$ $\sim$ 6.5$\%$ over DSSD. And the AP of our detector (59.4$\%$) is better than most of state-of-the-art detectors on PASCAL VOC, such as the RefineDet320 (54.7$\%$), DIoU (55.8$\%$), DAFS (58.7$\%$), Cascade R-CNN (58.5$\%$). When the input size is increased to 512, our method can achieve 61.2$\%$ AP which is comparable to most of state-of-the-art detectors at the same scale.
\subsection{Ablation Study}
In order to demonstrate the effectiveness of different components in our TPNet, we construct the ablation experiments on PASCAL VOC. As shown in Table~\ref{sample-table2}, we mainly analyze the effectiveness of the following components: T block, F block and RIoU loss. For a fair comparison, all models are trained on VOC 2007 \texttt{trainval} + VOC 2012 \texttt{trainval} and tested on VOC 2007 \texttt{test} with the input size of 320. The stricter COCO-style Average Precision (averaged AP at IoUs from 0.5 to 0.9 with an interval of 0.1) is also adopted in our ablation experiments.
\begin{table*}
\vspace{-0.1cm}
  \caption{Ablation results of each component (i.e.,T block, F block and RIoU loss) in TPNet, in which ResNet-50 is adopted as the backbone.}
  \small
  \vspace{-0.2cm}
  \label{sample-table2}
  \centering
  \begin{tabular}{cc|p{0.9cm}p{0.9cm}p{0.9cm}p{0.9cm}p{0.9cm}p{0.9cm}}

    \toprule

     \multicolumn{2}{c|}{method }              &  AP& AP$_{50}$ & AP$_{60}$ & AP$_{70}$ & AP$_{80}$ & AP$_{90}$   \\
    \hline
     \multicolumn{2}{c|}{Baseline(DSSD) }      &56.1 & 79.6  &74.8&64.1&46.1&16.0   \\
     \multicolumn{2}{c|}{Baseline+T block}   &57.6&80.6&76.1&65.0&47.3&19.0\\
     \multicolumn{2}{c|}{Baseline+T block+F block} &57.9&80.6&76.2&65.4&48.0&19.2\\
     \multicolumn{2}{c|}{Baseline+T block+F block+RIoU loss }  & 59.4&80.3&76.3&66.8&50.9&22.5\\
    \bottomrule
  \end{tabular}
  \vspace{-0.2cm}
\end{table*}

\textbf{Two-Pronged Transductive (T) Blocks and Fusion (F) Block.} We first conduct the ablation experiments to verify the effectiveness of the T block and F block. As shown in Table~\ref{sample-table2}, we can find that the detector can achieve 57.6$\%$ AP when we only add the T blocks to the base network, which improves the AP by 1.5$\%$ (from 56.1$\%$ to 57.6$\%$). The reason is that the T blocks really improve the performance of the classification task in low-level layers and regression task in high-level layers by exploring the relationship and two-pronged effect among different layers in both backward and forward directions. And when we add the Fusion Block to the network, the AP can be slightly improved (from 57.6$\%$ to 57.9$\%$). In addition, we also conduct experiments to analyze the effects of the two parts in T blocks, i.e., Multi-neighbor backward transfer (T-MBT) and forward transfer (T-FT). As shown in Table~\ref{sample-table3}, the detector can improve the AP by 0.7$\%$ (from 56.1$\%$ to 56.8$\%$) when we only adopt the T-MBT in T blocks. And when we add the T-FT into the T blocks, our detector can further improve AP by 0.8$\%$ (from 56.8$\%$ to 57.6$\%$). These experiments demonstrate that both T-MBT and T-FT play an important role in the network.
\begin{table}
  \caption{The effectiveness of the Multi-neighbor backward transfer (T-MBT) and forward transfer (T-FT) on VOC 2007 \texttt{test} set.}
  \small
  \vspace{-0.3cm}
  \label{sample-table3}
  \centering
  \begin{tabular}{cc|p{0.5cm}p{0.5cm}p{0.5cm}p{0.5cm}p{0.5cm}p{0.5cm}}

    \toprule

     \multicolumn{2}{c|}{T block }              &  AP& AP$_{50}$ & AP$_{60}$ & AP$_{70}$ & AP$_{80}$ & AP$_{90}$   \\
    \hline
    \multicolumn{2}{c|}{Baseline}      &56.1 & 79.6  &74.8&64.1&46.1&16.0   \\
    \multicolumn{2}{c|}{Baseline+T-MBT}   &56.8&80.3&75.5&64.1&46.8&17.4\\
     \multicolumn{2}{c|}{Baseline+T-MBT+T-FT}   &57.6&80.6&76.1&65.0&47.3&19.0\\
    \bottomrule
  \end{tabular}
\vspace{-0.5cm}
\end{table}

\begin{table}
  \caption{Performance analysis of the inflection point $\beta$ of the gradient on VOC 2007 \texttt{test} set.}
  \small
  \vspace{-0.3cm}
  \label{sample-table4}
  \centering
  \begin{tabular}{p{0.8cm}|p{0.8cm}p{0.8cm}p{0.8cm}p{0.8cm}p{0.8cm}p{0.8cm}}

    \toprule

     \ \ $\beta$               &  AP& AP$_{50}$ & AP$_{60}$ & AP$_{70}$ & AP$_{80}$ & AP$_{90}$   \\
    \hline
     \ 0.85      &58.7&80.2&76.5&66.3&50.0&20.7\\
     \ 0.90   &59.1&80.2&76.3&66.6&50.5&21.7\\
     \ 0.95   & 59.4&80.3&76.3&66.8&50.9&22.5\\
     \ 1.0   &59.2&80.2&76.2&66.4&50.7&22.4\\
    \bottomrule
  \end{tabular}
\end{table}

\begin{table}
  \caption{The effectiveness of training original SSD detector with different IoU based localization losses, ResNet-50-FPN is adopted as the backbone. }
  \small
  \vspace{-0.3cm}
  \label{sample-table5}
  \centering
  \begin{tabular}{cc|p{0.5cm}p{0.5cm}p{0.5cm}p{0.5cm}p{0.5cm}p{0.5cm}}

    \toprule

     \multicolumn{2}{c|}{Location loss }              &  AP& AP$_{50}$ & AP$_{60}$ & AP$_{70}$ & AP$_{80}$ & AP$_{90}$   \\
    \hline
     \multicolumn{2}{c|}{SSD+IoU loss }        &54.9&78.1   &73.4&62.9&45.6&14.6   \\
     \multicolumn{2}{c|}{SSD+GIoU loss}   &55.3&78.4&74.1 &63.5&45.9&14.6 \\
     \multicolumn{2}{c|}{SSD+DIoU loss}   &55.8&78.9&74.6 &64.0&46.2&15.5 \\
     \multicolumn{2}{c|}{\textbf{SSD+RIoU loss}(Ours)}   &\textbf{56.8}&\textbf{79.2}&\textbf{73.6}&\textbf{64.8}&\textbf{48.1}&\textbf{18.4}\\
    \bottomrule
  \end{tabular}
  \vspace{-0.4cm}
\end{table}

\begin{table*}
  \caption{Results on MS COCO \texttt{test-dev} set. The $\dagger$ means multi-scale inference.}
  \vspace{-0.1cm}
  \label{sample-table6}
  \centering
  \begin{tabular}{c|c|c|ccc|ccc}
    \toprule
    Method           &   Backbone      & FPS&AP & AP$_{50}$ & AP$_{75}$ &  AP$_{S}$& AP$_{M}$ &AP$_{L}$ \\

    \hline
    \footnotesize\emph{two-stage:}&  &&&&&&&\\
    Faster R-CNN \cite{ren2015faster}             & VGG-16 & 7 &21.9 & 42.7& -&-&-& -   \\
    Libra R-CNN \cite{pang2019libra}        &ResNet-101-FPN&6.8 &40.3 &61.3 &43.9 &22.9 &43.1 &51.0  \\
    TridentNet     \cite{li2019scale}       &ResNet-101&2.7& 42.7 &63.6 &46.5& 23.9& 46.6& 56.6  \\
    \hline
    \hline
    \footnotesize\emph{one-stage:}&  &&&&&&&\\
    SSD300 \cite{liu2016ssd}             & VGG-16  & 43&25.1 & 43.1&25.8& 6.6 & 25.9 & 41.4    \\
    YOLOv2 \cite{redmon2017yolo9000}              & Darknet-19    &  40  &21.6 & 44.0& 19.2&  5.0 & 22.4 & 35.5  \\
    DSSD321 \cite{fu2017dssd}            & ResNet-101   &  9.5  &28.0 & 46.1& 29.2&  7.4 & 28.1 & 47.6   \\
    RefineDet320 \cite{zhang2018single}  & ResNet-101&-& 32.0 &51.4& 34.2 &10.5& 34.7 &50.4                  \\
    DAFS320 \cite{li2019dynamic}        & ResNet-101&- &33.2 &52.7 &35.7 &10.9& 35.1 &\textbf{52.0}  \\
    \textbf{TPNet320}(Ours)              & ResNet-101  & 25.7 & \textbf{34.2} & \textbf{53.1}& \textbf{36.4}&\textbf{13.6} & \textbf{36.8} &50.5  \\
    \hline
    SSD512 \cite{liu2016ssd}           & VGG-16   &22  & 28.8 &48.5& 30.3&10.9 & 31.8 & 43.5   \\
    DSSD513 \cite{fu2017dssd}            & ResNet-101  & 5.5    & 33.2 & 53.3& 35.2&13.0 & 35.4 & 51.1  \\
    RefineDet512 \cite{zhang2018single} &ResNet-101 &-&36.4 &57.5 &39.5 &16.6 &39.9 &51.4                    \\
    DAFS512 \cite{li2019dynamic}       & ResNet101 &-&38.6 &58.9 &42.2 &17.2 &42.2 &54.8   \\
    RetinaNet800 \cite{lin2017focal}            & ResNet-101-FPN &  5   &39.1 &59.1 &42.3 &21.8 &42.7 &50.2  \\
    GHM-C + GHM-R \cite{li2019gradient}     & ResNet-101-FPN   &  4.8   &39.9 &\textbf{60.8} &42.5 &20.3 &43.6 &54.1\\
    CornerNet \cite{law2018cornernet} &Hourglass-104&4.4 &40.5 &56.5 &43.1 &19.4 &42.7 &53.9                \\
    \textbf{TPNet512}(Ours)             & ResNet-101 & 13.9  & 39.6 & 58.5&  42.8&20.5& 45.3 & 53.3      \\
    \textbf{TPNet512}$\dagger$ (Ours)           & ResNet-101  & - & \textbf{41.2} &  59.9 &\textbf{44.2} & \textbf{22.6} & \textbf{46.3}& \ \textbf{55.0 }     \\
    \bottomrule
  \end{tabular}
  \vspace{-0.2cm}
\end{table*}
\textbf{Performance analysis of our Rectified IoU loss.} In order to validate the effectiveness of the RIoU loss, we conduct three ablation experiments. Firstly, we use the RIoU loss instead of smooth $\mathcal{L}_1$ loss to train our TPNet. As shown in Table~\ref{sample-table2}, our TPNet trained with RIoU loss can improve AP by 1.5$\%$ compared with the detector trained with smooth $\mathcal{L}_1$ loss (from 57.9$\%$ to 59.4$\%$). Especially, the performance is largely improved by 3$\%$ at higher IoU threshold (i.e., 0.8 and 0.9), which demonstrates that our RIoU loss can substantially improve the model localization accuracy. Secondly, we conduct experiments to analyze the effect of the inflection point of the gradient $\beta$. Results are presented in Table~\ref{sample-table4}, in which four different inflection points of the gradient ranging from 0.85 to 1.0 are experimented. The best AP (59.4$\%$) of RIoU is obtained when the $\beta$ is 0.95. And the AP is slightly decreased when the $\beta=1.0$. This proves that it is unreasonable to up-weight the gradients of examples when the regression is perfect ($IoU \rightarrow 1$), as is described in section~\ref{RIoU_loss}. Thirdly, in order to validate the compatibility and generality of RIoU loss for object detection, we also conduct experiments on the original SSD and compare our RIoU loss with some other IoU based localization losses. The results are presented in Table~\ref{sample-table5}, from which we can see that the AP of the SSD with RIoU loss (56.8$\%$) is better than SSD with other IoU based localization loss, such as IoU loss (54.9$\%$), GIoU  loss (55.3$\%$) and DIoU loss (55.8$\%$).
\vspace{-0.0cm}
\subsection{Experiments on MS COCO}
To further validate our method, we also evaluate our TPNet on MS COCO 2017. Following the protocol in MS COCO, we use the \texttt{train} set (118, 287 images) for training and the \texttt{test-dev} set (20, 288 images) for evaluation. By submitting the detection result to \texttt{test-dev} evaluation server, we can download the final evaluation result.

We compare the evaluation results of our TPNet on MS COCO \texttt{test-dev} set with the results of some state-of-the-art detectors in Table~\ref{sample-table6}. ResNet-101 is adopted as the backbone in our experiments. Without bells and whistles, when the input size is 320, our TPNet with ResNet-101 produces 34.2$\%$ mAP that is better than most of one-stage detectors. Our model outperforms the baseline DSSD320 by 6.2$\%$ mAP (from 28.0$\%$ to 34.2$\%$). When the input size is increased to 512, our method can achieve 39.6$\%$ mAP with ResNet-101, which is comparable to most of state-of-the-art detectors at the same scale. Especially, our detector can achieve 41.2$\%$ mAP when we adopt the multi-scale inference strategy, which is competitive to most of one-stage detectors.
\section{Conclusion}
In this paper, we propose a novel accurate detector (TPNet), in which two novel contributions are included. First, we introduce a novel two-pronged transductive idea to explore the relationship and two-pronged effect among different layers for both backward and forward feature transfer. Under the guidance of the two-pronged idea, the proposed TPNet consists of multiple effective transductive (T) blocks. The T blocks achieve the bidirectional enhancement of features by multi-neighbor backward transfer and forward transfer. A relation has been established between the high-level and low-level layers of the network, which enables them to help each other, promote each other, and acquire more discriminative features. Second, in order to prevent the gradients of localization loss from being dominated by outliers during training phase and ensure the bounding box regression ability of the whole detector, we introduce a new localization loss, named RIoU loss, which can up-weight the gradients of examples with high IoU while suppressing the gradients of examples with low IoU. We carry out benchmark experiments on the PASCAL VOC and MS COCO datasets and the results demonstrate the state-of-the-art detection performance of our TPNet. The proposed T block and RIoU loss can be plugged and played in existing detectors, which will be our future work.
\begin{acks}
This work was supported by the National Science Fund of China under Grants (61771079) and Chongqing Youth Talent Program.
\end{acks}


\bibliographystyle{ACM-Reference-Format}
\balance 
\bibliography{sample-base}

\end{document}